# A New Pairwise Deep Learning Feature For Environmental Microorganism Image Analysis

Frank Kulwa[a], Chen Li[a,1], Jinghua Zhang[a], Kimiaki Shirahama[b], Sergey Kosov[c], Xin Zhao[a], Hongzan Sun[d], Tao Jiang[e], Marcin Grzegorzek[f]

[a] *Microscopic Image and Medical Image Analysis Group, College of Medicine and Biological Information Engineering, Northeastern University, Shenyang 110169, PR China*
[b] *Faculty of Science and Engineering, Kindai University, Kindai, Japan*
[c] *Faculty of Data Engineering, Jacobs University Bremen, Bremen, Germany*
[d] *Shengjing Hospital of China Medical University, Shenyang, China*
[e] *Control Engineering College, Chengdu University of Information Technology, Chengdu, China*
[f] *Institute for Medical Informatics, University of Lubeck, Ratzeburger Allee 160, 23538 Lubeck, Germany*

**Abstract**

*Environmental microorganism* (EM) offers a high-efficient, harmless, and low-cost solution to environmental pollution. They are used in sanitation, monitoring, and decomposition of environmental pollutants. However, this depends on the proper identification of suitable microorganisms. In order to fasten, low the cost, increase consistency and accuracy of identification, we propose the novel pairwise deep learning features to analyze microorganisms. The pairwise deep learning features technique combines the capability of handcrafted and deep learning features. In this technique we, leverage the Shi and Tomasi interest points by extracting deep learning features from patches which are centered at interest points' locations. Then, to increase the number of potential features that have intermediate spatial characteristics between nearby interest points, we use Delaunay triangulation theorem and straight-line geometric theorem to pair the nearby deep learning features. The potential of pairwise features is justified on the classification of EMs using SVMs, *k*-NN, and Random Forest classifier. The pairwise features obtain outstanding results of 99.17%, 91.34%, 91.32%, 91.48%, and 99.56%, which are the increase of about 5.95%, 62.40%, 62.37%, 61.84%, and 3.23% in accuracy, F1-score, recall, precision, and specificity respectively, compared to non-paired deep learning features.

[1] Corresponding author: Chen Li
*Email address:* lichen201096@hotmail.com (Chen Li )

## 1. Introduction

Microorganisms are tiny living organisms which can appear as unicellular, multicellular, and acellular [1]. Some are advantageous, and some are harmful to human health and environments. For instance, *Lactobacteria* are beneficial microorganisms, which decompose substances to give nutrients to plants [2], *Mmicrothrix parvicella* causes bulking in activated sludge [3], and *Epistylis* is used as an indicator of poor water quality. Examples of harmful microorganisms are microbacteria tuberculosis that causes tuberculosis and Rift Valley virus which causes Rift Valley fever. Deep understanding of microorganisms helps to leverage their beneficial side and eradicate the negative effects of harmful ones. Microorganisms are found almost everywhere, in unclean fluids, air, soil, and as parasites in other organisms such as humans and animals. Most of them have physical characteristics (for example, color, shape, texture, and size) that cannot be easily distinguished by naked eyes. In order to enhance analysis, microscopes are used to magnify the view of microorganisms. Then identification and classification of microorganisms are done by the expert (operator) based on observed features. Although the use of operator is real-time and flexible to changes of the examined samples, it is a slow process, expensive, tedious, biased to expert skills, inconsistent, and the results depend on operator's daily moods. To overcome such drawbacks, we introduce the new pairwise deep learning features, which can be used for automatic image classification of microorganism images. The proposed technique for extracting pairwise features is capable of highlighting more objects of interest (microorganisms) on the image, reducing the influence of the background, and associating noise on the image during classification. Thus, the features extracted are more robust and represent distinctive characteristics of the microorganism without the need for pre-processing, such as image denoising and segmentation. These features have shown a dramatic improvement in the classification of microorganisms. The workflow of the proposed technique is shown in figure 1 below.

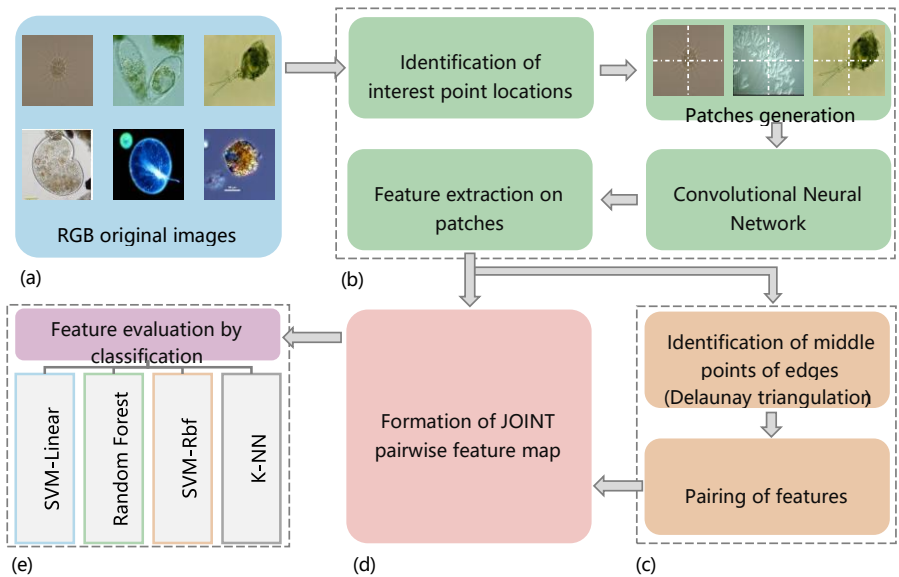

Figure 1: The workflow of the proposed method.

The workflow shown in figure 1 from (a) to (e) is demonstrated below.
(a) RGB original images: This is the original microorganism dataset whose features are to be extracted. (b) Feature extraction: In this step, we extract deep learning features from patches which are centered at interest points. (c) Feature pairing: In this section, the extracted features are paired using the Delaunay triangulation theorem and the middle point of the straight-line theorem. (d) Joint pairwise feature maps formation: The paired features and original features (from interest points) are combined to form a joint pairwise feature for each image. (e) Application of the pairwise features: At this stage, the potential joint pairwise features can be used for the classification task. In this experiment, some of the prominent classifiers are used to test the potential of the paired features for classification tasks. Such classifiers are support vector machine (SVM), random forest, and *k*-nearest neighbors (*k*-NN).

To test the potential of the new pairwise features, we perform the image classification of *Environmental Microorganisms* (EMs). Some of the EMs images are shown in figure 2.

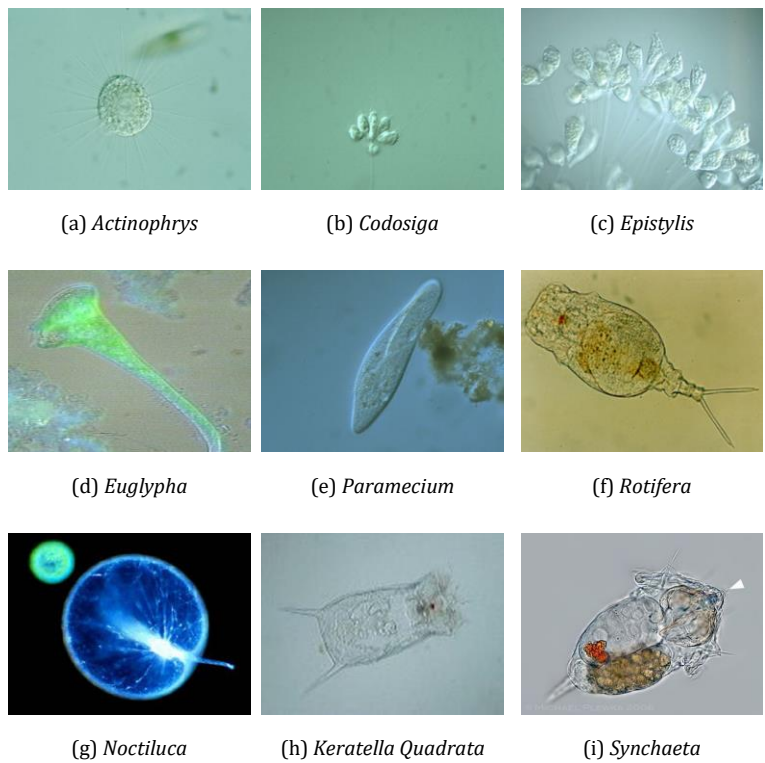

(a) *Actinophrys*  (b) *Codosiga*  (c) *Epistylis*

(d) *Euglypha*  (e) *Paramecium*  (f) *Rotifera*

(g) *Noctiluca*  (h) *Keratella Quadrata*  (i) *Synchaeta*

Figure 2: Some examples of EM images.

The contributions of this paper can be folded as described below:

1. We combine and leverage the capability of interest points (handcrafted features) and deep learning features by extracting deep learning features on small image patches of size 40 × 40, which are centred at corner interest points' locations. The interest points are determined using Shi and Tomasi theorem. This enables the location of distinctive points on the image where local distinctive deep learning features of an object (microorganism) can be extracted.

2.      Speculating that the middle point between two nearby patches (interest points) have intermediate spatial features, we pair the feature maps of two nearby interest points. This increases the percentage of potential distinctive features of an object. The pairing is achieved using the Delaunay triangular theorem, which concentrates the triangles inside the object of interest, thus highlights more the object of interest against the noisy background.

This paper is organized as follows; Section 2 reviews in detail recent related works. Section 3 describes our proposed methods and different key points of our contributions in depth, while experiment and analysis of the results are discussed in section 4. Finally, the conclusion and future work are highlighted in section 5.

**2. Related Work**

Several works have been done to automate the analysis of microorganisms using image processing techniques. Image processing involves six basic steps, which are image acquisition, pre-processing (denoising, segmentation, enhancement, etc.), feature extraction, post-processing (feature selection, feature fusion, feature vector reduction, etc.), classifier design (classification), and evaluation [4]. Feature extraction is an important stage in image analysis. Good features result in good classification results. For instance, in the [5], the Fourier descriptors are used as features for the classification of TB-bacilli from sputum smear using multilayer perceptron neural network, and it achieves outstanding results of 92.5%. In [6], random forest (RF) and SVM are used for the classification of soil microorganism using potential colorimetric features and histogram features, where SVM achieves an accuracy of 98.19% and RF achieves 98.41%. This shows that the good features lead to good results in many classifiers. Generally, there are two types of feature extraction techniques, handcrafted and feature learning [7], as indicated in table 1.

Handcrafted features are manual features that are extracted based on prior knowledge. For example, color, texture, geometric features (area, perimeter, and length), global shape, and local shape (such as SIFT and SURF). Interest points are very useful in finding handcrafted features, particularly local features (SIFT and SURF). Interest points are distinct regions/points used to differentiate between one object (image) to another [16]. Examples of interest points are corner, blob, and ridge descriptors. They play a significant role in image matching and classification tasks. For instance, in [8], [9] SIFT features (which are extracted from corner descriptors of 10 channels of different color modes) are used for image matching of EMs. In [10], edge and Fourier descriptors are applied for the classification of EMs using SVM classifier. In [17], high accuracy results of 97.55% is achieved on the classification of TB bacilli using local SURF and location-oriented histogram features fed to the deep learning neural network. In [18], SIFT features are used as texture features in the classification of species of bacteria using SVMs and random forest classifiers. Interest points are useful in classification and image matching because they are invariant to rotation, translation, and changes in illumination. Moreover, the local image structure around the interest point is rich in local discriminant information content [19]. Thus, we take advantage of corner interest points' locations around the microorganism by extracting deep learning features around the interest points so as to capture the discriminant characteristics of the object of interest (microorganism) on the image.

Table 1: Categories of feature extraction techniques.

| Categories | Specific feature (techniques) examples | Related works |
|---|---|---|
| Hand crafted | Geometric features (Area, perimeter), Local features (SIFT, SURF), Color, Texture | [8], [9], [10], [11] |
| Feature Learning | Deep learning (VGG-16, AlexNet, ResNet), BoVW | [12], [13], [14], [15] |

Feature learning (features) is a high dimension feature formed by the composition of local features such as SIFT. Examples of feature learning are the bag of visual words (BoVW) [20], sparse coding (which analyses a large number of images to learn a set of bases where each expresses a characteristics pattern of a patch [21]), and deep learning features. Deep learning features are extracted by deep learning convolutional networks such as ResNet, AlexNet, and VGG-

16. Deep learning networks represent high-level features composed of low-level ones. They have superior descriptive power than handcrafted features methods [22], because they mimic the effectiveness of visual cortex in human brain for feature extraction [23]. Thus, they find many applications in microorganism image classification tasks. For example, in our previous [24] a pre-trained VGG-16 is exploited for pixel-level deep learning feature extraction from EM images, which are used for classification using a CRF model. In [25], a unary classification method is proposed to detect bacteria in different environmental conditions, where a convolutional deep belief network and a SVM classifier are used to segment the object of interest, then a six layers convolutional neural network (CNN) is applied for feature extraction and classification. In [26] a 13 layers CNN is used in extraction of color channel features from plankton images. Then these features are used for classification into 38 related taxonomic groups of plankton. The SparseConvNet is used for classification. In [27] a R-CNN is used for deep feature learning and classification of cyanobacteria images into five classes.

VGG-16 is one of the most useful and powerful models in image classification tasks due to its high feature learning capability. For example, in [12], VGG-16 achieves an outstanding performance on detection and classification of bacterial and viral *pneumonia* from x-ray images. In [28], a pre-trained VGG-16 is used in the classification of *Radiolarians*, which are biostratigraphic indicators for palaeogeographic reconstructions. To leverage the superiority of VGG-16, in this paper, we use it in the extraction of deep learning features at each detected corner descriptor's location (patch).

Bag of visual words (BoVW) is among the most popular feature learning techniques because of its robustness and simplicity. However, due to the orderless representation of local features, it does not give optimal performance. To remedy that, some works have considered spatial arrangement of features to discover higher order structures for improving the performance of BoVW in object matching and classification works [29, 30]. Among the techniques of arranging features is by pairing of spatial close visual words [31]. For instance, in [15] and [32], pairing is done on visual words (where feature descriptors are mapped to the visual words before pairing, then pairing is done on visual words). However, the pairing of visual words seems to ignore the underlying distribution of pairs of nearby local feature descriptors. As a remedy to that, [13] and [14] propose the pairing of spatial close local feature descriptors (SIFT) and treat them as data points in a joint feature space before the building of BoVW. In this way, more improvement is achieved in the object recognition performance on the tested challenging dataset. Inspired by the idea of pairing features, and to the best of our knowledge, no work has been done on the pairing of deep learning features for the classification of microorganism images, thus in this paper, we pair deep learning features extracted from corner descriptors' locations and use them for classification of EMs.

**3. Methods**

The novel techniques for extracting and pairing of features are elaborated in detail in this section. Moreover, some of the intermediate experimental results are shown in this section. The main focus is to increase the potential of the deep learning features for the classification of microorganisms. The description follows the workflow of figure 1.

*3.1. Features Extraction*

*3.1.1. Shi and Tomasi Intest Points' Location*

Although many images in the EM dataset have low contrast and transparency challenges, the boundary and corner physical characteristics can easily be detected compared to other interest points. Therefore, we choose to use corner interest points. A corner is defined as the location (point) in the image where a slight shift in the location will lead to a large change in intensity in both horizontal (X) and vertical (Y) axes. It can also be defined as the intersection of points on objects contour edges, which retain important feature information of the objects [33]. Shi and Tomasi corners theorem is among the most superior corner theorems [34]. Simply the Shi and Tomasi theorem operates on three
steps:

Firstly, it is to determine the window, which produces high variation in intensity with a small move in the $X$ and $Y$-axis. Numerically, to determine a window that can produce large variation, let the window be centred at $(x, y)$ and the intensity at this point be $I(x, y)$. $I(x, y)$ is an individual intensity at a position, which can have a value from 0 to 255 for gray level image. When the window is shifted by $(u, v)$, the intensity at the new location will be $I(x + u, y + v)$ and $[I(x + u, y + v) - I(x, y)]$ is the difference in intensity due to shift. For a corner, this difference must be high. Therefore, we maximize this term by differentiating it with respect to $x$ and $y$. Let $w(x, y)$ be the weights of pixels over the rectangular or a Gaussian window, Then, $E(u, v)$ which is the difference between the original and the shifted window, is defined as:

$$E(u, v) = \sum_{x,y} w(x, y)[I(x + u, y + v) - (x, y)]^2 \tag{1}$$

Apply the Taylor series with only the first order, which is

$$T(x, y) = f(u, v) + (x - u)f_x(u, v) + (y - v)f_y(u, v) \tag{2}$$

Rewrite the shifted intensity using the above formula:

$$I(x + u, y + v) = I(x, y) + \frac{d(x,y)}{dx}(u) + \frac{d(x,y)}{dy}(v) \tag{3}$$

Let $\frac{d(x,y)}{dx} = I_x$ and $\frac{d(x,y)}{dy} = I_y$, then $I_x$ and $I_y$ are image derivatives in $X$ and $Y$ directions, respectively. So, $E(u, v) = \sum_{x,y} w(x, y)[I(x, y) + I_x u + I_y v - I(x, y)]^2$. Furthermore,

$$E(u, v) = \sum_{x,y} w(x, y)[I_x u + I_y v]^2 \tag{4}$$

Expand the above equation as:

$$E(u,v) = \sum_{x,y} w(x,y)[I_x^2 u^2 + I_y^2 v^2 + 2I_x I_y uv] \qquad (5)$$

Take $u, v$ out and rewriting in matrix notation, the equation becomes:

$$E(u,v) = (u,v) M \begin{pmatrix} u \\ v \end{pmatrix} \qquad (6)$$

where,

$$M = w(x,y) \begin{pmatrix} \sum_{x,y} I_x^2 & \sum_{x,y} I_x I_y \\ \sum_{x,y} I_x I_y & \sum_{x,y} I_y^2 \end{pmatrix}$$

where $M$ is a symmetric $2x2$ matrix whose eigenvalues are used to determine whether the scanned window contains a corner.

Secondly, Calculating the score value $S$ associated with scanned window [34]. It is given by;

$$S = min(\lambda_1, \lambda_2) \qquad (7)$$

where $\lambda_1$ and $\lambda_2$ are eigenvalues of the matrix $M$.

The third step is finding the points along the shift of the window, which can be regarded as corners. For the point to be regarded as the corner, the score value $S$ should be greater than the specified value (if both the $\lambda_1$ and $\lambda_2$ are greater than the minimum threshold values).

Shi and Tomasi theorem show superiority by having stability, invariant to scale changes, translation, and rotation [34]. Moreover, comparing with Harris corner points which we applied in our previous work [9], Shi and Tomasi gives better results and more useful interest points than Harris'. Thus, we use it to determine corner points on every image. Examples of images with corner points indicated on them are shown in figure 3.

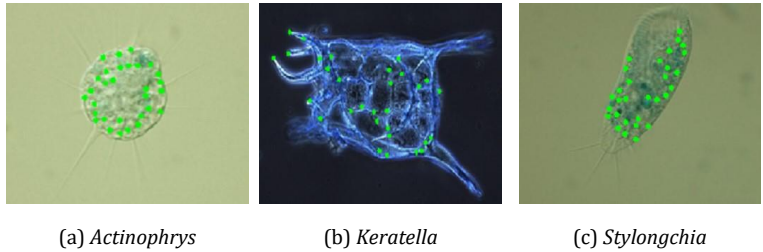

| (a) *Actinophrys* | (b) *Keratella* | (c) *Stylongchia* |

Figure 3: Detected Shi and Tomasi corner points (in green) on EMs.

As can be seen from figure 3, the interest points are capable of detecting corner points which are rich in distinct information about the object of interest (microorganism) and differentiate it from the background. In the practical case, we limit the number of corner points between 10 to 15 (due to the deep model's computational complexity during the extraction of features). Then the coordinates of each corner point are identified and stored. We take advantage of the corner points by meshing each image into patches of size 40×40, which are centered at each corner point as shown in figure 5 part (a) and (b). Then from each patch, we extract deep learning features using convolution neural network VGG-16.

*3.1.2. VGG-16*

VGG-16 is a very deep convolution neural network for image recognition, proposed by Simonyan et al. in [35]. It is upgraded from AlexNet by replacing large sized kernel filters (11 and 5) with 3 × 3. It has achieved high accuracy in many image classification tasks. It contains 21 layers with only 16 weight layers, which include 13 convolution layers with very small receptive fields of 3×3 (which gives its capability to capture the pattern of tiny information fields), followed by max-pooling layers of size 2 × 2 and stride 2, which decreases the spatial resolution of the feature maps. In the end, there are three fully connected layers, which combine all learned features from previous layers and generalize them for classification. ReLu activation function is applied to all hidden layers. Lastly is the classifier layer. Because VGG-16 is a typical and famous CNN with high effectiveness, we choose it as the baseline CNN model to design our pairwise method. To leverage the fully connected (FC) layers, we extract deep learning features on the last FC layer. The dimension of each extracted feature is about 1 × 1000 dimensions. The figure 4 shows the VGG-16 network layers and the point from which deep learning features are extracted.

Due to the small number of EMs which cannot train the VGG-16 from scratch for better results, we use the transfer learning concept to optimize the VGG-16 extracted features. VGG-16 network, pre-trained on the ImageNet

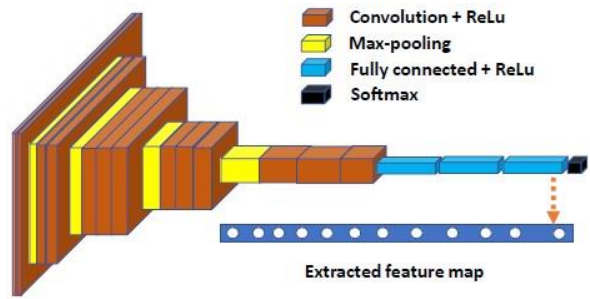

Figure 4: VGG-16 network showing feature extraction layer (fully connected layer 3).

dataset has proven success in many works when fine-turned on other datasets for classification [36]. Therefore, we fine-tune the pre-trained VGG-16 using the EMs and extract deep learning features. For each image, 10 patches of size

40×40 are meshed out and from each patch deep learning features are extracted (each patch is centered at interest points' coordinate). Then 10 features for each image are stored parallel to their corresponding interest points' coordinates. Figure 5 summarizes the process of deep learning features extraction.

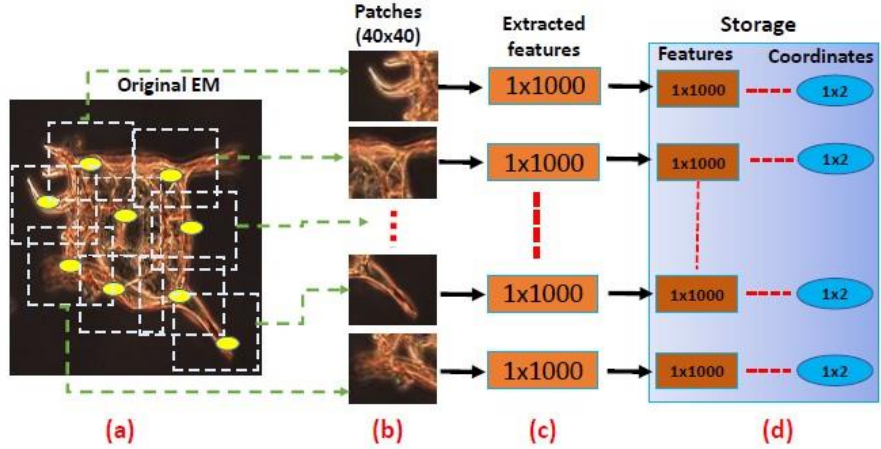

Figure 5: The work flow for extraction of deep learning features from EM images (a) detection of interest points' coordinates indicated by yellow colour on the original image, (b) meshing of patches which are centred at interest points, (c) deep learning feature extraction, (d) storage of features and interest points' coordinates.

## *3.2. Feature Pairing: Delaunay Triangulation (DT) Theorem and Middle Point of the Straight Line Theorem*

To pair feature maps that have been extracted from the interest points' coordinates, we use the Delaunay triangulation (DT) theorem. DT theorem is one of the most robust graphical theorems for the representation of data. It is the triangulation theorem which forms triangles (Delaunay triangles) by connecting each data (coordinates) to its nearest neighbor, such that the circumcircle associated with each triangle does not contain a point in its interior [37]. Geometrically, Delaunay triangulation for a given set **A** of discrete data in a plane is a triangulation (DT), such that no data in **A** is inside the circumcircle of any triangle in DT(**A**). Delaunay triangulation maximizes the minimum angle of all the angles of the triangles in the triangulation [38]. It is very effective for presentation of scattered data as it concentrates all data inside the major circumcircle formed by the most outer triangle as shown in figure 6 (b). Due to strong presentation power, it is used in many image matching works [39, 40]. Moreover, it is tolerable to spatial displacement of data (image objects) because it keeps the same association of the nearest objects within the image, regarded that the distortion is uniform all over the image.

The Delaunay triangle edges are formed by connecting nearest neighbor data points. This means two points (vertices) which share the same edge (line) have close related characteristics (features). Thus, the middle point of the edge contains features which are an average of the edge end point features. Although (from our experiments) few middle points might be out of the EM's body, which will have non similar characteristics between the edge end points; these points are very few (less than 5% of all the middle points). More than 95% of the middle points are within the main body of the EM (foreground) and have intermediate characteristics between the corresponding edge end points, as it can be observed in figure 6 (c). Owing to this, we pair the features that correlate to the vertices sharing the same edge to get the features of the middle point of edges. In this way, we increase the number of potential distinctive features for the microorganism, as shown in figure 6 (c). The pairing of features is done using the geometric theorem of the middle point of the straight line, because the edges of the triangles are straight lines. This is done by averaging the two feature vectors (maps) corresponding to each edge end coordinates as described in the equation 8 and 9. The edge coordinates are the interest points' coordinates with their corresponding feature vectors (1 × 1000 dimensions) extracted from patches.

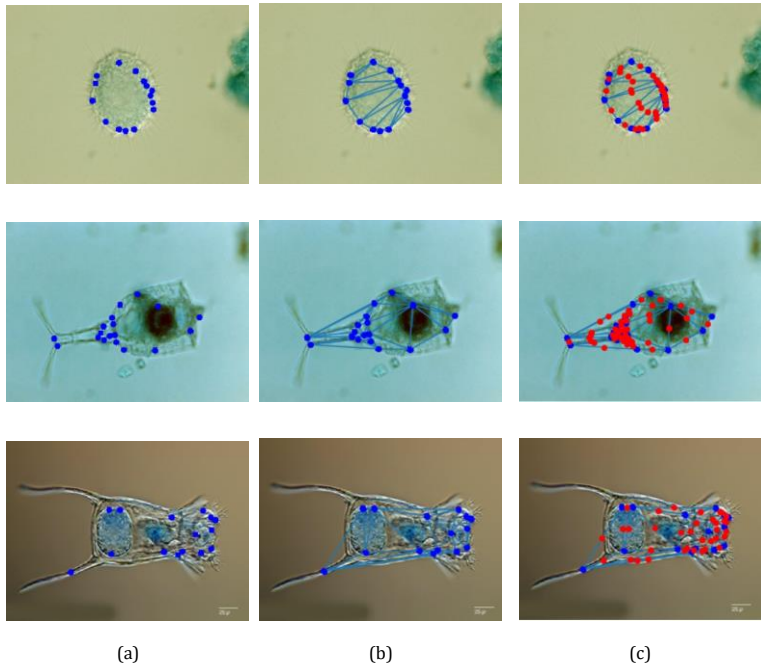

(a)           (b)           (c)

Figure 6: Pairing of features. (a) Detection of interest point coordinates which are indicated in blue colour, (b) application of Delaunay triangulation, (c) pairing of features at middle points of edges which are indicated in red. Blue coloured points are interest points.

Let the coordinate of the two end points of an edge be represented by $(X_1, Y_1)$ and $(X_2, Y_2)$. The corresponding feature maps of the patches centered at these two points be $F_1$ and $F_2$. The middle point coordinate $X_m, Y_m$ is given by:

$$X_m, Y_m = \frac{(X_1 + X_2)}{2}, \frac{(Y_1 + Y_2)}{2} \tag{8}$$

The pairwise feature map $F_m$, which corresponds to middle point $X_m, Y_m$ is given by:

$$F_m = \frac{(F_1 + F_2)}{2} \tag{9}$$

On average, 36 to 43 pairwise features (*Fm*) are formed from 10 original features for each image.

*3.3. Joint Pairwise Feature Formation*

At this stage, we join the features formed on the interest points' coordinates (*F*1, *F*2...) and pairwise features (*Fm*...). The average amount of pairwise features for each image is between 36 and 43. 10 features originate from interest points. Thus, we form the joint feature maps by appending these features vertically. This joining style has shown the best results from the tests done during experiments. The average joint feature maps sizes range from 46×1000 to 53×1000 for different images. Therefore, each joint feature map corresponds to one original image. Because the dominant features are pairwise features, we name the features as joint pairwise features (Pairwise deep learning features). To justify the potential of the pairwise deep learning features, we use them for the classification of EMs.

## 4. Experiments and Analysis

*4.1. Experimental Settings*

*4.1.1. Dataset*

In this experiment, we use Environmental Microorganism Dataset 5th Version (EMDS-5) for testing the potential of the pairwise features for the classification task. This dataset is comprised of 21 classes. Each class contains 20 images of the same species. These species include *Actinophrys, Arcella, Aspidisca Codosiga Colpoda Epistylis Euglypha Paramecium Rotifera Vorticella Noctiluca Ceratium Stentor, Siprostomum, Keratella Quadrala, Euglena, Gymnodinium, Gonyaulax, Phacus, Stylongchia*, and *Synchaeta*. Some of the EMs images are shown in figure 2. For presentation purposes we name these classes as class 1 (c1), class 2 (c2), class 3 (c3)...class 20 (c20), class 21 (c21) respectively. In total, there 420 available images. True corners of the foreground are the most important in this research. To reduce the possibility of false corners, we crop all images that have outer highlighted square frames at the edges of the images and remain with only the true background and foreground. Then all images are resized to 480 × 360 pixel sizes.

*4.1.2. Experimental Environments*

To conduct the experiments, we use a workstation with Intel (R) Core (TM) i7-7700 CPU with a speed of 3.60Hz. RAM of 32GB and NVIDIA GeForceGTX 1080 8GB. For implementation of the networks, we use python 3.6, Keras framework and Tensorflow 1.7 as the backend.

*4.1.3. Classifiers*

Five classifiers (from skilearn library) are used in this experiment to justify the usefulness of the paired deep learning features for the classification of microorganisms. These classifiers are support vector machine with the linear kernel (SVM-Linear), support vector machine with radial basis function kernel (SVM-rbf), $k$-nearest neighbors ($k$-NN), and ensemble based random forest classifier. In the SVM-Linear, the regularization is set to 1, the iteration count is not limited, and the numerical tolerance used is 0.001 to optimize the classifier. In the SVM-rbf, the decision function is set to "one vs. one" (ovo). In *k*-NN the optimal *k* is found to be 21, points are weighted based on their inverse distance, and the algorithm used to compute the nearest neighbor is kd-tree. The random forest classifiers are trained with 200 trees, and the maximum depth is set to 21.

*4.2. Evaluation Metrics*

In order to evaluate and compare the classification results quantitatively, we use accuracy (Acc), F1-score, Recall, Precision (Prec), and Specificity (Spec). They are the most standard metrics for the evaluation of classification performance. Table 2, summarizes the definition of these metrics.

Table 2: Definitions of used metrics

| Metric | Definition | Metric | Definition |
|---|---|---|---|
| Acc, Spec | $\frac{TP+TN}{TP+FP+TN+FN}, \frac{TN}{TN+FP}$ | F1-Score | $\frac{2TP}{2TP+FP+FN}$ |
| Recall | $\frac{TP}{TP+FN}$ | Prec | $\frac{TP}{TP+FP}$ |

From table 2, true positive (TP) is the number of accurately predicted positive samples, true negative (TN) is the number of correctly classified negative samples, the number of negative samples classified as positive is false positive (FP), and the number of positive instances predicted as negative is a false negative (FN).

*4.3. Evaluation of the Non-paired Features From Patches*

Firstly, we mesh each image into patches which are centered at Shi Tomasi interest points. The size of each patch is 40 × 40 pixel sizes. Then each patch is reshaped to 224 × 224, which is the input size for the VGG-16 network. On average, each microorganism image results in 10 patches. Finally, deep learning features (feature vector) are extracted from each patch. Each feature vector is having a dimension of 1 × 1000 size. Therefore, for each image 10 deep learning features are extracted. During the pilot experiments, we tested two options of combining the feature maps extracted from the same image. The first option is by appending all features horizontally to form a high dimension feature vector of 1 × 10000 which represents one image. The second option is to treat each feature independently. This is similar to appending the features vertically resulting into 10×1000-dimension matrix. In this option, each feature from the same image points to the same label. To validate which option gives better performance. These features are used for binary image classification of class 1 and 2 of EMs. The Linear SVM is used for classification. The accuracy results obtained are 55.00% and 77.50% for the horizontal pending option and vertical option, respectively. Then, we opt to use the vertical appending when performing the classification of 21 classes. Therefore, each image results in a feature matrix of 10 × 1000 dimensions.

During image classification of 21 EMs classes, the features are randomly shuffled and partitioned at the ratio of 1:1 for training and testing sets. Then classification of the EMs is performed into 21 classes. In order to reduce the bias of feature selection during shuffling, the whole process of shuffling randomly, partitioning, training, and testing is repeated 50 times. Each time the classification results (confusion matrices) are recorded. The average score for all 50 iterations for each classifier are presented in figure 7 below:

Figure7 shows that all classifiers have achieved high accuracy and specificity while the F1-score, precision, and Recall are still very low. This implies that the extracted features from interest points are potential features. However, the number of features representing the positive class (TP) during classification is less than the rest (TN) classes. To address this, we increase the number of potential features by pairing them, which will reduce the asymmetric of true class (TP) against the rest (TN) during classification.

*4.4. Evaluation of Pairwise Deep Learning Features*

In order to increase the number of potential features, the pairing of features is performed. During pairing, each image is represented by an average of 46 features resulted from pairing. These features are vertically appended. All features within the same image points to the similar label. Then classification of

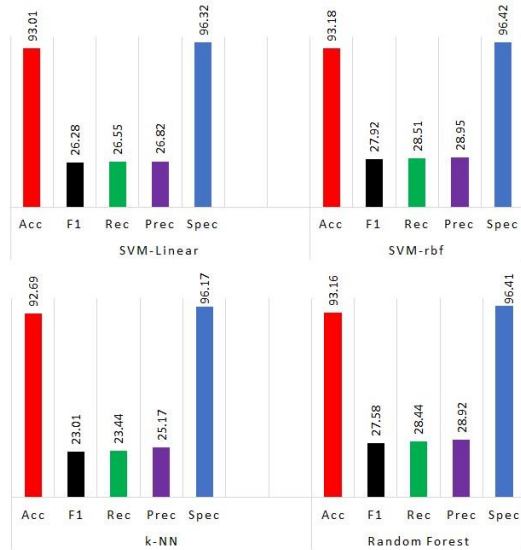

Figure 7: The average classification scores of 21 classes when using the non-paired deep learning features extracted from interest points (patches) only. The results are presented in percentage.

21 classes is performed to justify the potential of the pairwise features. Similarly, in order to reduce the bias of some features on the classification results, cross validation is applied, where data are randomly shuffled then divided into the ratio of 1:1 for training and testing and classification is done using the four classifiers, then the process from shuffling to classification is repeated 50 times for each classifier. Each time classification results and confusion matrices are recorded. The average results for all iterations are shown in figure 8.

From the results shown in figure 8, the pairwise features have significantly improved the classification results. Comparing with the average results of nonpaired deep learning features in figure 7, the improvement of about 6.08%, 64.23%, 63.93%, 63.89%, and 3.20% in accuracy, F1-score, recall, precision, and specificity, respectively, is observed on SVM-Linear classifier, 2.13%, 22.84%, 22.22%, 23.50%, and 1.12% in accuracy, F1-score, recall, precision, and specificity, respectively, is observed on SVM-Rbf classifier, 4.63%, 49.00%, 48.47%, 47.62%, and 2.42% in accuracy, F1-score, recall, precision, and specificity, respectively, is observed on k-NN classifier, 4.69%, 49.93%, 49.03%, 49.26%, and 2.46% in accuracy, F1-score, recall, precision, and specificity, respectively, is observed on Random Forest classifier.

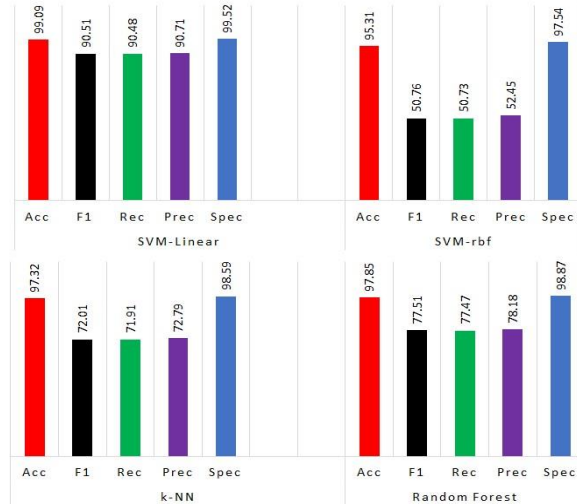

Figure 8: The average classification scores of 21 classes when using the pairwise deep learning features. The results are presented in percentage

*4.5. Comparison of Pairwise and Non-pairwise Deep Learning Features*

To validate the performance of the pairwise deep learning features more and make sure the average results in sections 4.3 and 4.4 are not affected by outliers, we observe and compare the maximum values of the classification results shown in figure 9.

Firstly, observing the results in figure 9 and figures 7, 8, the highest difference between the average and maximum results on pairwise features related graphs is 1.75% F1-Score observed on the SVM-rbf while most of the other differences are less than 1.00%. The highest difference on non-paired features results is 2.66% F1-score observed on SVM-Linear. These small values of difference show that the average results are not influenced by any outlier values.

Secondly, we compare the maximum results of the non-paired and paired features as observed in figure 9. A significant improvement of about 5.95%, 62.40%, 62.37%, 61.84%, and 3.23% in accuracy, F1-score, recall, precision, and specificity respectively, is observed on SVM-Linear classifier, 2.14%, 22.86%, 22.36%, 23.58%, and 1.11% in accuracy, F1-score, recall, precision and specificity, respectively, is observed on SVM-rbf classifier. 4.60%, 48.58%, 47.76%, 46.70%, and 2.41% in accuracy, F1-score, recall, precision and specificity, respectively, is observed on *k*-NN classifier and 4.57%, 48.44%, 48.05%, 47.79%, and 2.40% in accuracy, F1-score, recall, precision and specificity, respectively, is observed on Random Forest classifier.

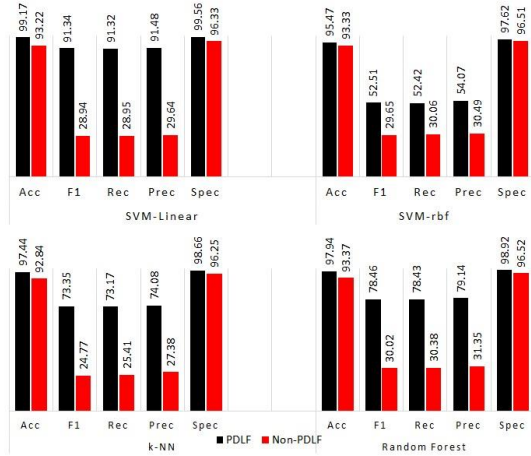

Figure 9: The maximum classification scores of 21 classes when using the non-pairwise deep learning features (Non-PDLF) extracted from interest points (patches) only which are represented with red bars and the pairwise deep learning features (PDLF) which are represented in black. All the results are presented in percentage.

*4.6. Training and Testing Time Evaluation*

The average training and testing time for each classifier when using the non-paired features and paired features are shown in table 3.

As observed from table 3, the maximum time requirement for using pairwise features to perform the classification of 21 EM classes is 135.15 seconds (about 2.25 minutes) for training and 183.67 seconds (about 3.05 minutes) for testing, while others are less than that. Although the time requirement for the pairwise features is higher than for non-paired features, they are generally very low and feasible for practical applications of classification tasks.

Table 3: Training and testing time when using non-paired deep learning features (Non-PDLF) from interest points only and when using pairwise deep learning features (PDLF).

|  |  | SVM-Linear | SVM-rbf | k-NN | RF |
|---|---|---|---|---|---|
| Train (Sec) | Non-PDLF | 8.02 | 7.81 | 0.5 | 13.04 |
|  | PDLF | 135.15 | 122.58 | 4.26 | 62.31 |
| Test (Sec) | Non-PDLF | 5.25 | 5.69 | 10.15 | 0.5 |
|  | PDLF | 74.97 | 94.64 | 183.67 | 2.5 |

*4.7. Discussion*

To analyze the performance of pairwise features on each class, we observe the confusion matrices for each classifier in figure 10.

Observing the matrices in figure 10, the pairwise deep learning features have shown outstanding classification performance in almost all classes on all classifiers. This is because of the high potential of the pairwise features in capturing the distinct spatial characteristics of the microorganism images. However, in most classifier results (SVM-rbf, *k*-NN and Random Forest), the least performance is observed on class 10 *vorticella* (c10). This is because our method has experienced challenges in extracting enough useful features on class 10. The biggest challenge on the class 10 is lack of distinct corner points on the microorganism (extremely low contrast between the microorganisms and background). As indicated in figure 11.

Although the corner detection method used is superior and has overcome challenges in almost 20 classes, the extremely low contrast which leads to lack of notable intensity difference between the background and foreground pixels on the majority of individual images of class 10, has resulted in false corner detection (as shown in blue points in figure 11 outside the microorganism body), which leads to false pairwise features extracted outside the microorganism's body. This is the main course of misclassification experienced in class 10 by almost all classifiers except SVM-Linear. Generally, the application of the best corner detection technique influences the results/potential of the pairwise features.

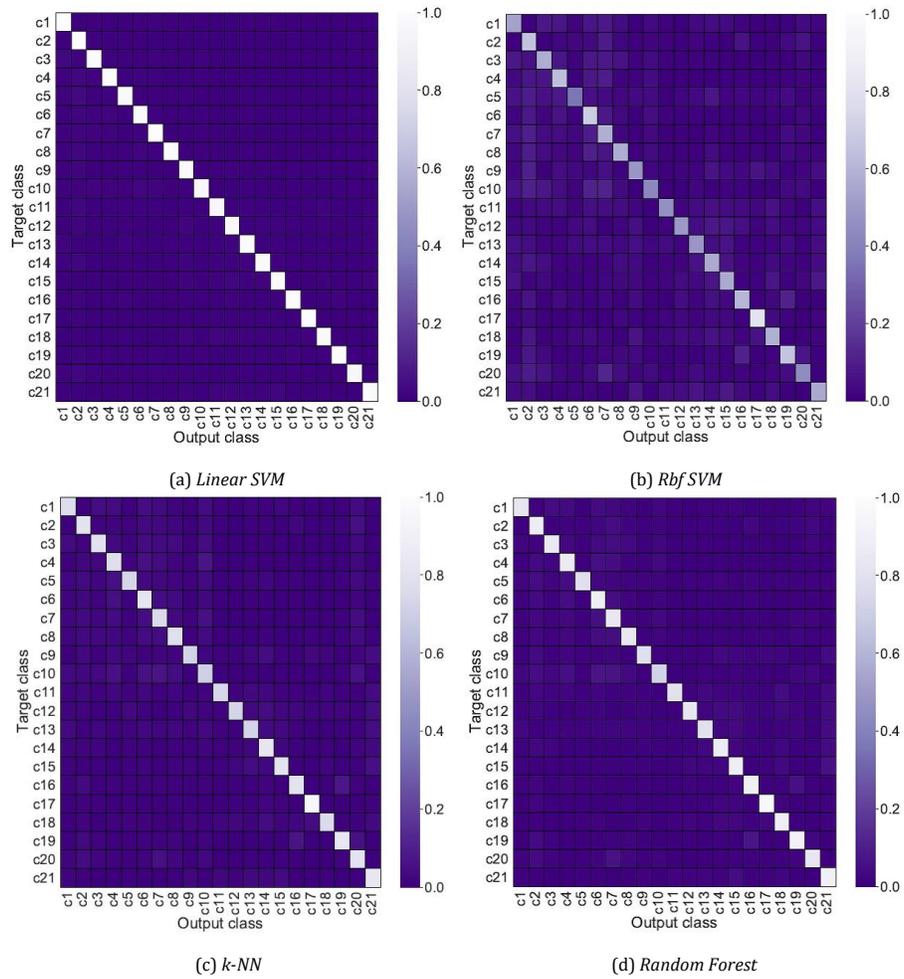

(a) *Linear SVM*          (b) *Rbf SVM*

(c) *k-NN*          (d) *Random Forest*

Figure 10: Average confusion matrices showing the classification performance of (a) Linear SVM, (b) SVM which rbf kernel, (c) *k*-NN and (d) Random Forest when using pairwise deep learning features.

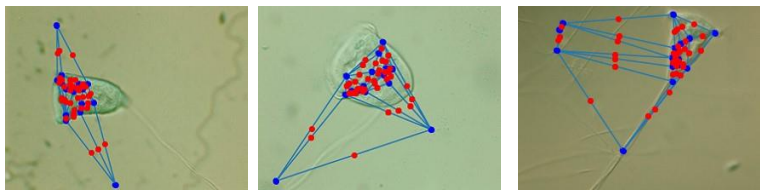

Figure 11: The False corners (false non-pairwise features) which are indicated in blue outside the microorganism body and false paired features which are indicated in red outside the microorganism body of class 10 (c10).

## 5. Conclusion and Future Work

In this paper, the novel pairwise deep learning features are proposed. To increase the potential of deep learning features, a combination of handcrafted and deep learning is applied. Due to the high capacity of interest points, we use Shi and Tomasi to represent distinct points on images where deep learning features are extracted. Then pairing is done between the nearby features based on interest points locations. The pairwise features prove their potential by obtaining the highest classification results on EMs images in all used classifiers. This technique in not only suitable for identification of EM images but also in medical related microorganism images such Plasmodium from blood sample images and TB bacteria from stained sputum samples. It can also find use in facial recognition tasks. Moreover, the patch-based deep learning features (without pairing) can be suitable in the classification of tumors such as brain and breast tumors using MRI or Ultrasound images when features are extracted from blob interest points locations. In future works, we will extract and pair the deep learning features from other prominent models such as ResNet50 and InceptionV3 and use them to perform the classification and segmentation of microorganisms.

**Acknowledgements**

This work is supported by the "Natural Science Foundation of China" (No. 61806047). We thank Prof. Beihai Zhou, Dr. Fangshu Ma from the University of Science and Technology Beijing, PR China, and Prof. Yanling Zou from Freiburg University, Germany, for their previous cooperation in this work. We thank Miss Zixian Li and Mr. Guoxian Li for their important discussion. We also thank B.E. Xuemin Zhu from Johns Hopkins University, US and B.E. Bolin Lu from Huazhong University of Science and Technology, PR China, for their careful work in the EMDS-5 image data preparation.